\documentclass[conference]{IEEEtran}
\IEEEoverridecommandlockouts
\usepackage{cite}
\usepackage{amsmath,amssymb,amsfonts}
\usepackage{algorithmic}
\usepackage{graphicx}
\usepackage{textcomp}
\usepackage{xcolor}
\def\BibTeX{{\rm B\kern-.05em{\sc i\kern-.025em b}\kern-.08em
    T\kern-.1667em\lower.7ex\hbox{E}\kern-.125emX}}
\begin{document}
\IEEEoverridecommandlockouts
\IEEEpubid{\begin{minipage}[t]{\textwidth}\ \\[10pt]
        \centering\normalsize{978-1-6654-0945-2/21/\$31.00 \copyright 2021 IEEE}
\end{minipage}}

\title{Electromyography Signal Classification Using Deep Learning\\
}

\author{\IEEEauthorblockN{Mekia Shigute Gaso}
\IEEEauthorblockA{\textit{Department of Computer Science} \\
\textit{Ala-Too International University}\\
Bishkek, Kyrgyzstan \\
mekiashigute.gaso@alatoo.edu.kg}
\and
\IEEEauthorblockN{Selcuk Cankurt}
\IEEEauthorblockA{\textit{Department of Computer Science} \\
\textit{Suleyman Demirel University}\\
 Kaskelen, Kazakhstan \\
\IEEEauthorblockA{\textit{Department of Computer Science} \\
\textit{Ala-Too International University}\\
Bishkek, Kyrgyzstan \\
s.cankurt@iaau.edu.kg}}
\and
\IEEEauthorblockN{Abdulhamit Subasi}
\IEEEauthorblockA{\textit{Institute of Biomedicine} \\
\textit{University of Turku}\\
Turku, Finland \\
abdulhamit.subasi@utu.fi}
}

\maketitle

\begin{abstract}
We have implemented a deep learning model with L2 regularization and trained it on Electromyography (EMG) data. The data comprises of EMG signals collected from control group, myopathy and ALS patients. Our proposed deep neural network consists of  eight  layers;  five fully connected, two batch normalization and one  dropout layers. The data is divided into training and testing sections by subsequently dividing the training data into sub-training and validation sections. Having implemented this model, an accuracy of 99 percent is achieved on the test data set. The model was able to distinguishes the normal cases (control group) from the others at a precision  of  100 percent  and  classify  the  myopathy  and  ALS with  high  accuracy  of 97.4 and  98.2 percents, respectively.  Thus we believe that, this highly improved classification accuracies will be beneficial for their use in the clinical diagnosis of neuromuscular disorders.
\end{abstract}

\begin{IEEEkeywords}
EMG, deep learning, DNN, EMG signal classification, Myopathy, ALS.
\end{IEEEkeywords}

\section{Introduction}
What is commonly refereed to as Electromyography (EMG) is a process of obtaining and registering electrical signals from the musculoskeletal system in human body. Signals obtained in these ways can be used for various medical applications. Some of these applications can generally be grouped into those which are related to clinical and to those which focus on the interactions of humans with computer systems. All of these signals require to carefully decompose, processes and analyse the signal obtained in various intricate procedures to finally classify them into those which are showing medical disorders related to muscular functions and those which are healthy subjects \cite{c1}.

The physical procedure of acquiring EMG signal involves penetrating a very narrow needle which contains an electrode into the subject's skin, and proceeding to record the signals as the muscles are contracting and relaxing following some movement \cite{c2}. These signals are observed to be often very chaotic and to be strongly tied up to the intrinsic properties of the muscles themselves as they are related to their structures and their functionalities. As a result, these signals usually exhibit quite noisy, complex, and high-dimensional characteristics \cite{c4}. 

In the medical procedure when the need arise the physician may proceed to take the EMG signal of his subject. A muscle showing weakness, lack of flexibility, inflicting pain during movement are some of the symptoms that will help a physician to propose these procedure of acquiring EMG signals from his patients. During these processes it is expected that the electrodes will send a very small amount of electrical signals to the patients nerves so that the later responds to it. These data can then be transformed using computer programs into figures and numerical values that can later be interpreted by the physician \cite{c2}.

EMG signals are affected in the processes of acquisition of the signals from the reaction emerging from the muscles in what is commonly known as the environmental noise. In addition to those there are effects emanating from the intrinsic structure of the muscle. These are often related to the interaction of the induced electric current with the muscle as the signal crosses different layers of the muscles as it penetrates them. The electrical changes resulted from the interaction of the neuromuscular system of the muscles with that of the electrodes which is inserted in the muscles is commonly represented in what is known as the motor unit action potential (MUAP). The MUAP provides vital information in investigating if a subject is likely to have developed neuromuscular disorders. Therefore, all these collective factors add up to the complex structure that is retained in the final signals that is acquired, thereby introducing for a need to clean or/and de-noise the signal (see for example the discussions given in \cite{gokgoz} and references therein). 

Resolving the above addressed issues in order to distil the most important information from EMG data requires developing various strategies for de-noising the signal. Together with these methods, often an intermediate steps are introduced in processing the EMG data which are known as, the feature extraction methods. These methods have proved themselves when combined with de-noising techniques of the EMG data to have shown significantly higher classification accuracies, in order to classify the signals of subjects showing neuromuscular disorders from those who are not affected. For instance, E. Gokgoz \& A. Subasi in there work \cite{gokgoz} have reported that using discrete wavelet transform (DWT) as a feature extraction technique has increased the accuracy of the EMG classification. Following there suggestions we will also apply DWT in our work which is a method by which signals are split up into multiple frequency ranges.

%%%%%%%%%%%%%%%%%%%%%%%%%%%%%%%%########### ends here ####################################%%%%%%%%%%%%%%%%%%

\section{Review of related literature}

In the following we will briefly discuss some closely related research works with our study reported in this paper. It is common practice to record EMG signals either with a surface electrodes or with a needle electrodes. By using the later case in acquiring the EMG signal, multidimensional multiscale parser (MMP) has been employed for encoding electromyographic signals in the work of \cite{eddie2008emg}. They have conducted experiments with a real signals which was acquired in laboratory and there by showing the technique they have used is a robust one when compared with its counterparts. They have reported the evaluated percent root mean of encoding EMG signal which has the False Positives of 91.1\%  and those which are correctly identified of 95\%.

EMG signal need to be decomposed so that information is extracted from it that can be used for clinical purpose. During the decomposition process what is known as  motor unit potential trains (MUPTs) is obtained the validity of which need to be ascertained as error may be resulted in the process \cite{Parsaei2011}. Having employed various techniques for this purpose they author of \cite{Parsaei2011} have compared the outputs on simulated as well as real data. They have reported that the adaptive gap-based Duda and Hart (AGDH) method they applied have shown an improved accuracies of about 91.3\% and 94.7\% in classifying accurately the simulated as well as the real data, respectively. 

Similar study has been carried out in assessing the validity of MUPTs in reference \cite{Parsaei2010}. The MUPTs are acquired having decomposed an EMG signal which in the first place is obtained through a needle-detected EMG signal. The authors have proposed two techniques to find out about the validity of the MUPTs based on the shape information decoded in the motor unit potential (MUP). Both of the proposed methods make use of the gap statistic and jump algorithms which are applied to find some statistical properties of the data such as finding the number of groups in the data sets. Among the two methods the gap statistics has excelled to have generated an accuracy of $\sim$92\% and $\sim$94\% on the simulated and real data, respectively.

The author of reference \cite{Li2013} have emphasized on the decomposition of the EMG signals into which they are primarily composed of, namely the motor unit (MU) firing times and action potential shapes, as these can be applied for investigating neuromuscular disorders involving various clinical researches. They introduced multi-channel decomposition algorithms known as Montreal and Fuzzy Expert on both simulated as well as experimental data. They analysed the performance of the two algorithms on the data which is acquired from twelve subjects of different age categories at different level of contraction. Performance levels were cross checked for showing resemblance between the two algorithms for both the experimental data there by comparing an accuracy predicted priory for that of the simulated data. The median agreement of the result between the two algorithms have been shown to be $\sim$96\% accurate.

Density-based method can be used to automatically decompose a single-channel intramuscular EMG signals into their compositions trains, the MUAP's \cite{Marateb2011}. In this method, outliers that belongs to the superpositions and factitious potentials are recognized  and were able to be excluded in advance of the classification process.  Then the MUAP templates are pointed out by an adaptive density-based clustering procedures. These decomposition techniques have been applied on signals acquired from lower level contractions of a maximum of 30\% on various types of muscles. Having compared their result with an expert manual decomposition, they report that their algorithm have identified 80\% of the total $\sim$230 motor unit trains with $>$ 90\% accuracy.

EMG decomposition system called EMGTool which is believed to be adaptive and multifaceted tool for detailed EMG analysis has been introduced by \cite{Nikolic2010}. This tool has a capacity to extract the constituent MUAPs and firing patterns (FPs) which can be used for numerical analysis from the EMG signal acquired with a minimal effort for clinical diagnostics. The authors have shown these in this work, there by successfully excluding  critical parameters with a fixed threshold by turning them into an adaptive ones. The algorithm they employed for the EMG signal decomposition have 3 stages, which are dedicated for that of segmentation, clustering, and resolution of compound segments. Having introduced different techniques for validating the results, an accuracy of $\sim$95\% have been obtained to accurately identify the FPs.

There is a growing interest from those classification schemes which makes use of machine learning to apply decision tree algorithms in order to classify biomedical signals. The later are often required to be thoroughly de-noised and followed by the introduction of quite  efficient feature extraction technique to finally obtain higher level of accuracies. For instance, E. Gokgoz \& A. Subasi in there work of \cite{gokgoz} has classified the acquired EMG signals by (i) employing some de-noising techniques for cleaning the signals, (ii) applying some feature extraction techniques for processing the data in order to obtain some of its essential features, (iii) and finally have used some classifiers on both testing and validating data sets. Thus, for de-noising the EMG signals they have used Multiscale Principal Component Analysis (MSPCA) which calculates the Principal Component Analysis (PCA) of the wavelet coefficients and combine the results at defined scales. Some of the essential feautures of the data have been extracted using Discrete Wavelet Transform (DWT) technique which is briefly discussed in section \ref{subsec:DWT}. Lastly, the classification is performed by using decision tree algorithms such as C4.5, CART and random forests \cite{gokgoz}. They have reported for the robustness of these design they used in its ability to automatically classify the EMG signals into those with neuromuscular disorder, i.e., myopathic, ALS or those which are normal. Having compared the results through various performance measures they report that the overall accuracy obtained in their work is $\sim$97\% accurate. Thus they have emphasised that their framework can seen to have the capability on the classification of EMG signals with a good accuracy.

Following similar procedure, we proposed a deep neural networks for classifying EMG data instead of machine learning techniques. We have used the same data sets as in the work of the authors of \cite{gokgoz}. We have introduced different  architecture which uses deep neural networks as given in Fig. \ref{fig1} (see section \ref{Sec: architecture} for the details) which has thus improved the accuracy. The contribution of our work is therefore to outline that the EMG signal classification accuracy has improved to be (99\%) from previous works (see for example \cite{gokgoz} for comparison) by making us of MSPCA de-noising methods and DWT feature extraction techniques. In addition to these, this is achieved because of our new implementation of the deep learning model that we will be presenting in the subsequent section.

The remaining parts of this paper is organized in the following way: in the next part (Section III), we provide the background section by introducing processing of the data with a through explanation of the feature extraction techniques used.  In Section IV the experimental data and the subjects along with the implementation of the deep learning model is explained. Section V provides the results obtained from our experiment with the use of various performance measures along with the confusion matrix for both the test and validation data sets followed by their corresponding discussions. In Section VI we provide the conclusion of our research work presented here.

%%%%%%%%%%%%%%%%%%%%%%%%%%%%%%%%%%%%%%%%%%%%%%%%%%%%%%%%%%%%%%%%%

\section{Background}

Here we will provide the background of our research having introduced the methods of data processing with a brief explanation of the feature extraction techniques used in this work.

\subsection{Multiscale Principal Component Analysis (MSPCA)}

Multiscale Principal Component Analysis (MSPCA) combines the characteristics and ability of Principal Components Analysis (PCA) to de-correlate the variables by obtaining firm inter-relationship.  Wavelet analysis is employed to find quite essential features and closely decorrelate the autocorrelated values. At each scale, MSPCA calculates the PCA of the wavelet coefficients and integrates the results at the marked scales. MSPCA is practicable for signal modeling which often shows changes across time interval and frequency ranges. Thus, this characteristics of MSPCA can be taken to be the advantage of the multiscale appraoch (see for example \cite{gokgoz} for the detailed description of this approach). In this study MSPCA is used for de-noising the raw EMG signals (see  \cite{gokgoz2014} for the detailed discussion of this topic).

%%%%%%%%%%%%%%%%%%%%%%%%%%%%%%%%%%%%%%%%%%%%%%%%%%%%%%%%%%%%%%%

\subsection{Discrete Wavelet Transform (DWT)}
\label{subsec:DWT}
Grossman and Morlet are the first ones in developing the mathematical framework describing the wavelet theory \cite{wavelet} in 1984, which allows for a method of describing a generic function of spatial and temporal varying scales by braking it down into its constituents. A decomposition of a signal with a better time resolution could be obtained by making use of the wavelet transform that splits up a signal into a set of basic functions which are known to as a wavelets. In general, wavelet-based techniques are applicable methods in analyzing various types of varying signals like EMG. For example discrete wavelet transform (DWT) can be used to measure and to shift the so called the mother wavelet. A discrete-time signal can also be disintegrated into a bunch of other smaller sets of signals by using DWT. The detailed description of this process is outlined in greater details in reference \cite{DWT}. Following the procedure explained in reference \cite{gokgoz}, a total of 27 features has been extracted which are used in this work whose details are given in section \ref{Sec: feature}.
 
%%%%%%%%%%%%%%%%%%%%%%%%%%%%%%%%%%%%%%%%%%%%%%%%%%%%%%%%%%%%%%%

\section{Implementation of the Deep Learning Model}
\label{sec:3}

%%%%%%%%%%%%%%%%%%%%%%%%%%%%%%%%%%%%%%%%%%%%%%%%%%%%%%%%%%%%%%%

\subsection{EMG Data}

Here we will give a brief description of the data sets which were used in our work. We have adopted the data sets, which were also used by E. Gokgoz \& A. Subasi, which is given in reference \cite{Nikolic2001}. The data is obtained from control group, myopathy and ALS patients. We briefly summarized the information about the group from which the EMG signal is acquired in a tabular form as in Table \ref{tab1}. The EMG signals were collected at low voluntary and constant level of contraction which is just above threshold by using concentric needle electrode. The signals are acquired from 5 spots located at varying location on the subjects body having inserted the needles at a different level of penetration.

\begin{table}[t]
\caption{General information regarding the three groups in which the subjects from which an EMG data is taken.}
\renewcommand{\arraystretch}{1.50}
\centering
 \begin{tabular}{|c | c | c | c |} 

 \hline
    \textbf{ Groups  and  some of} & \textbf{Control} & \textbf{Myopathy} & \textbf{ALS}  \\ [0.5ex] 
     \textbf{their attributes}    & \textbf{group }  & \textbf{group}    & \textbf{group} \\ 

 \hline
 \textbf{Number of subjects} &  10 & 7 & 8  \\ 
  \hline

 \textbf{Age category} & 21-37 & 19-63 & 35-67 \\ 
  \hline

 \textbf{Gender} & 4 male & 2 Male & 4 Male \\ 
  \textbf{distribution} & 6 Female & 5 Female & 4 Female \\ 

 \hline
 
  \textbf{Physical} & 6 very good & All & All \\ 
  \textbf{condition} & 3 good &  show &  show \\
  \textbf{of the subjects} & 1 bad & myopathy & ALS  \\

 \hline
 \end{tabular}
 
 %\hfill %to put some space between table and caption

 \label{tab1}
\end{table}

%%%%%%%%%%%%%%%%%%%%%%%%%%%%%%%%%%%%%%%%%%%%%%%%%%%%%%%%%%%%%%%

\subsection{Feature Extraction}
\label{Sec: feature}

EMG signals are segmented with a window length of 8192. Then we get 1200 instance for each of the three classes (i.e., ALS, Myopathy and Control group), which makes the length of the dataset 3600. 
In this study, the EMG signals are represented by using the following features of coefficients given in the formulas listed below, see also reference \cite{gokgoz}.

 \hspace{-0.34cm}(i) Mean of the coefficients for each sub-band: $$\rm{Mean} = \frac{\sum_{i=1}^n C_{i}}{n}.$$		(ii) Average power of the wavelet coefficients in each \\ $~~~~~$ sub-band: $$\rm{Average} = \frac{\sum_{i=0}^N (C_{i})^2}{N}.$$  (iii) Standard deviation of the coefficients in each sub-band: $$ \rm{Standard \ deviation} = \sqrt{ \frac{\sum_{i=1}^n (C_{i} - \mu )^2}{n}}.$$ (iv)	Ratio of mean values of neighbouring sub-bands: $$ \rm{Ratio} = \frac{\frac{\sum_{i=1}^n C_{i}}{n}}{\frac{\sum_{j=1}^n C_{j}}{n}}.$$ The 1$^{st}$ and the 2$^{nd}$ features are extracted for evaluation of the frequency distribution of the signal. The 3$^{rd}$ and the 4$^{th}$ features are extracted for the evaluation of the changes in the frequency distribution. Seven different features are extracted from (i), (ii) and (iii) each; and six different features are extracted from (iv). Hence a total of 27 features are extracted which consist of, the mean, the average power of the wavelet coefficients in each sub-band, the standard deviation, and the ratio of the mean values of neighboring sub-bands. These extracted features are used on the wavelet coefficients so as to make them more applicable and feasible. These features are calculated for D1–D6 and A6 frequency bands which were then used as input to the classifiers. Higher classification accuracy and lower computation cost is reported to be obtained when using “db4” wavelet filter \cite{gokgoz}. Finally, we have extracted a total of 27 features from the raw EMG signals, thus our dataset dimension became 3600 by 27.

%%%%%%%%%%%%%%%%%%%%%%%%%%%%%%%%%%%%%%%%%%%%%%%%%%%%%%%%%%%%%%%

\begin{figure}
\centering
\includegraphics[width=0.44\textwidth]{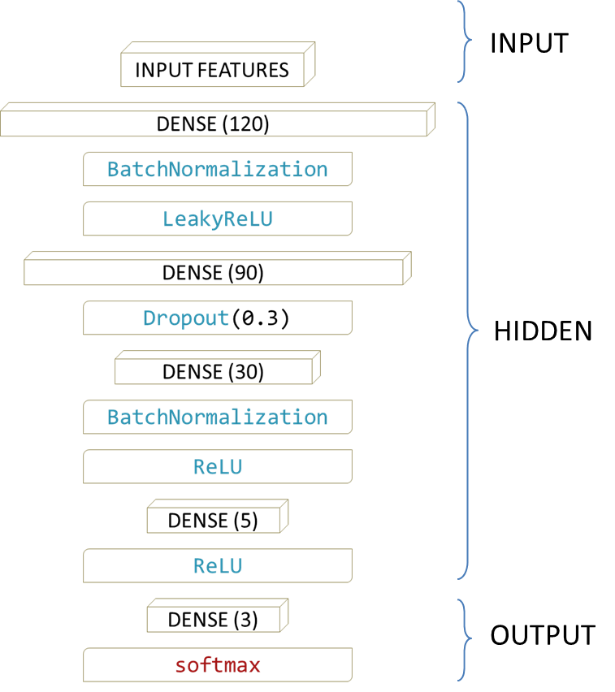} 
\caption{The architecture of the proposed deep neural network used in this work.}
\label{fig1}
\end{figure}

\subsection{Classification using Deep Neural Network}   
\label{Sec: architecture}
In this paper, we have implemented our own architecture of deep neural network with L2 regularization (weight decay) and trained it on EMG data. As depicted in Fig. \ref{fig1}, our proposed deep neural network contains eight layers; five fully connected, two batch normalization and one dropout layers. The overall design of the network architecture consists of three main sections: (1) An initial section which involves a single layer to present the feature vector of EMG signals to the network, (2) four stages of fully connected layers with different number of hidden neurons (120, 90, 30, 5 nodes), and (3) a final section of a fully connected layer with 3 nodes and softmax activation function. The last dense layer is used to combine the features which are used to classify the EMG signals. As a result, the number of the nodes in the last dense layer should match the number of classes in our data set. In this case, the output size is set to be 3 as in the number of the classes. 

The Leaky ReLU and the ReLU activation functions, which introduces the non-linearity to layers is employed to generate the activations of fully connected layers. The outputs of the fully connected layer are normalised with the use of softmax activation function. The former consists of positive numbers whose sum is equal to 1.0. These numbers thus entails information which can be used as the classification probabilities for the classification layer. This layer uses the probabilities returned by the softmax activation function for the corresponding input in order to assign the input to one of the disjoint classes and calculates the losses. We have trained the deep neural network using Adam (derived from adaptive moment estimation) optimization algorithm with a batch size of 150 samples and the cross-entropy loss function by shuffling the data at every epoch. 

%%%%%%%%%%%%%%%%%%%%%%%%%%%%%%%%%%%%%%%%%%%%%%%%%%%%%%%%%%%%%%%%%

\section{Results and Discussions}

\begin{figure}
\centering
\includegraphics[width=0.512\textwidth]{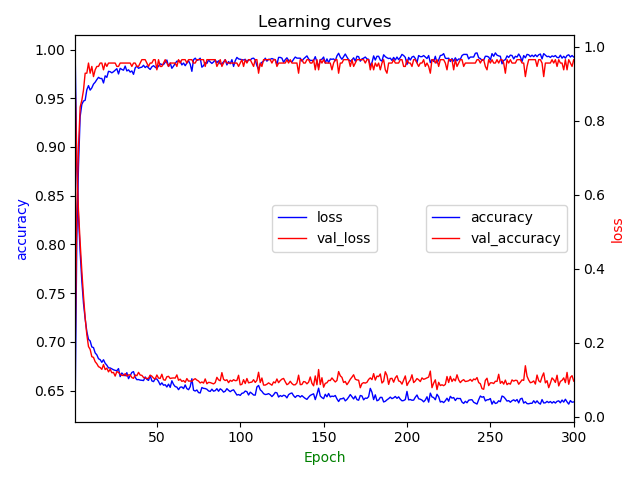} 
\caption{A graph of learning curves showing training progresses for various epochs with the losses and accuracies together with their corresponding accuracies of the validation data sets.}
\label{fig2}
\end{figure}

We have divided the data into training (80\%) and testing (20\%) sections. The training data is further divided into sub-training (90\%) and validation (10\%) sections. We have used, the sub-training data to train the network and update the weights, and the validation data to compute the accuracies at the regular intervals during training. The deep neural network's learning curves plotted in Fig. \ref{fig2} illustrate the training progress and show the mini-batch loss and accuracy together with the validation loss and accuracy. To estimate the final accuracy of the implemented deep learning model we have employed the test data. Our model has achieved 99\% accuracy on the test data set. 

\begin{figure}[b]
\centering
\includegraphics[width=0.53\textwidth]{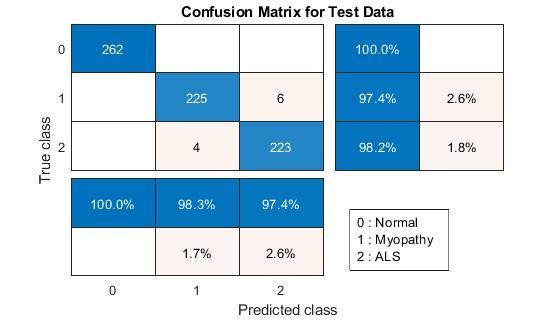} 
\caption{Confusion matrix for the test data set. The three classes namely, the normal, mypathy and ALS are represented by `0', `1' and `2', respectively.}
\label{fig3}
\end{figure}

Fig. \ref{fig3} displays the confusion matrix for the test data set by using column and row summaries. The trained deep neural network model distinguishes the normal cases from the others without confusing them (i.e., with the precision of 100\%) and classify the Myopathy and ALS with high accuracies of 97.4\% and 98.2\%, respectively. Training a deep network with eight layers is challenging as they can be sensitive to the initial configuration of the network and learning algorithm. Overfitting, which is simply memorization of the training data is a common problem for the traditional feed forward neural networks including the deep neural networks. This problem avoids the network to generalize the learning process and make accurate predictions for new data. Overfitting of a network can be monitored using the plot of learning curves during the training process. To reduce overfitting we have used, the L2 regularization at the first, second and third fully connected layers, two batch normalization layers after the first and third dense layers, and one dropout layer after the second dense layer. There are several weight regularization techniques, such as L1 and L2 regularizations, and each has a hyperparameter that must be configured. In reference \cite{c5} it has been reported that small amount of weight decay not only regularize the network but also improve the model’s accuracy. We have used the L2 of weight and bias regularization on the first, second and third dense layers, and carefully tuned the L2 regularization hyperparameter of weight decay to 0.000001. We have found that using this rate, instead of setting it to its default value of 0.01, is quite optimal in order to minimize the overfitting of the implemented model.

When the weights are updated after each mini-batch, the distribution of the inputs to layers may change, which is called internal covariate shift. Batch normalization standardizes the distribution of the inputs through the layers for each mini-batch \cite{c6}. In our design we have added batch normalization layers between the dense layers and Relu activations. We observed that adding batch normalization layer stabilizes the training process and improves the training accuracy. Dropout layer avoids the neurons from co-adapting extremely and prevents the overfitting \cite{c7}. Using both batch normalization and dropout techniques in the same dense layer is reported as not relevant \cite{c6}. Following  reference \cite{c6}, we have regularized the first and the third dense layers using batch normalization and the second dense layer using the dropout layer. 

The following are the computational environments that were used in this work. A deep-learning packages Tensorflow (v2.3.1) \cite{Abadi2015} back-end and Keras API (v2.4.0) \cite{Chollet2015} have been used to build deep neural networks. We have trained our proposed model on a PC system with Inter Core i5 9$^{th}$generation at 2.40 GHz CPU, GeForce GTX 1650 GPU and 16 GB RAM.

%%%%%%%%%%%%%%%%%%%%%%%%%%%%%%%%%%%%%%%%%%%%%%%%%%%%%%%%%%%%%%%

\section{Conclusion}

In this work we have implemented a deep learning model with L2  regularization   and   trained   it   on   EMG   data.   The   data comprises  of  signals  collected  from  control  group,  myopathy and ALS patients. We find a strong relationship between  our implementation of the deep neural network architecture,  and de-noising  and  decomposition  methods on the data, as results obtained in these work have revealed.  The contribution  of  this  study is therefore rests on  developing an effective and efficient EMG classification   algorithms   for   intramuscular   EMG   signals. Our proposed deep neural network contains eight layers; five fully connected, two batch normalization and one dropout layers. The data is divided into training and testing sections  by  subsequently  dividing  the  training  data  into  sub-training and validation sections. An accuracy of 99\% is achieved on  the  test  data  set.  The  model  was  able  to  distinguishes  the normal cases (control group) from the others at a precision of 100\% and classify the myopathy and ALS with high accuracy of  97.4\%  and  98.2\%,  respectively. Therefor, we believe that these highly improved classification accuracies will be beneficial for the clinical  diagnosis of neuromuscular disorders.

%%%%%%%%%%%%%%%%%%%%%%%%%%%%%%%%%%%%%%%%%%%%%%%%%%%%%%%%%%%%%%%%%

\section*{ACKNOWLEDGMENT}

M. S. G would like to give thanks to  Ala-Too International University for the scholarship opportunity which is given to her by Ala-Too International University, and Dr. S. Cankurt for the continuous guidance and for the wonderful supervision on her M.Sc. research and Dr. R. R. Mekuria for the fruitful discussions on some aspects of the writing up process of this research paper.

%%%%%%%%%%%%%%%%%%%%%%%%%%%%%%%%%%%%%%%%%%%%%%%%%%%%%%%%%%%%%%%%%


\begin{thebibliography}{99}


\bibitem{c1} Reaz, Mamun Bin Ibne, M Sazzad Hussain, and Faisal Mohd-Yasin (2006). “Tech-niques of EMG signal analysis: detection, processing, classification and applications”. In:Biological procedures online8.1, pp. 11–35.
 
\bibitem{c2} William Morrison M.D.  — Written by Danielle Moores — Updated on March 20,2018 Medically reviewed by (2018). Electromyography(EMG).URL:https://www.healthline.com/health/electromyography (visited on 05/24/2021).
 
%\bibitem{c3} Tuncer, Turker, Sengul Dogan, and Abdulhamit Subasi (2020). “Surface EMG signal classification using ternary pattern and discrete wavelet transform based feature extraction for hand movement recognition”. In:Biomedical Signal Processing and Control58, p. 101872.

\bibitem{c4} Lashgari, Elnaz and Uri Maoz (2020). “Electromyography Classification during Reach-to-Grasp Motion using Manifold Learning”. PLOS ONE doi: 10.1371/journal.pone.0255926.

\bibitem{gokgoz} E. Gokgoz, A. Subasi, Comparison of decision tree algorithms for EMG signal classification Biomedical Signal Processing and Control, 18 (2015), 138–144.


\bibitem{eddie2008emg} Eddie Filho, BL, Eduardo AB da Silva, Murilo B de Carvalho, (2008). “On EMG signal compression with recurrent patterns”. In:IEEE transactions on biomedicalengineering55.7, pp. 1920–1923.


\bibitem{Parsaei2011} Parsaei, Hossein, and Daniel W. Stashuk. "Adaptive motor unit potential train validation using MUP shape information." Medical engineering \& physics 33.5 (2011): 581-589.

 \bibitem{Parsaei2010} Parsaei,  Hossein  and  Daniel  W  Stashuk  (2010).  “Evaluating  a  motor  unit  potential train using cluster validation methods”. In:Canadian Student Conference onBiomedical Computing and Engineering, pp. 31–35


 \bibitem{Li2013} Li, Yejin et al. (2013). “Cross-comparison between two multi-channel EMG decomposition algorithms assessed with experimental and simulated data”.  In:201339th Annual Northeast Bioengineering Conference. IEEE, pp. 191–192


 \bibitem{Marateb2011} Marateb, H. R., Muceli, S., McGill, K. C., Merletti, R., \& Farina, D. (2011). “Robust decomposition of single-channel intramuscular EMG signals at low force levels”. In:Journal of neural engineering8.6, p. 066015


 \bibitem{Nikolic2010} Nikolic, Miki and Christian Krarup (2010). “EMGTools, an adaptive and versatiletool  for  detailed  EMG  analysis”.  In:IEEE  transactions  on  biomedical  engineering58.10, pp. 2707–2718.

\bibitem{Nikolic2001} Nikolic M. Detailed Analysis of Clinical Electromyography Signals EMG Decomposition, Findings and Firing Pattern Analysis in Controls and Patients with Myopathy and Amytrophic Lateral Sclerosis. PhD Thesis, Faculty of Health Science, University of Copenhagen, 2001. [The data are available as dataset N2001 at http://www.emglab.net]


\bibitem{gokgoz2014} Gokgoz E, Subasi A. Effect of multiscale PCA de-noising on EMG signal classification for diagnosis of neuromuscular disorders. J Med Syst. 2014 Apr;38(4):31. doi: 10.1007/s10916-014-0031-3. Epub 2014 Apr 3. PMID: 24696395.


\bibitem{wavelet} Grossmann, A. and J. Morlet, “Decomposition of Hardy functions into square integrable wavelets of constant shape”, SIAM J. Anal. 15, pp. 723-736, 1984.

\bibitem{DWT} Subasi, Abdulhamit, and Emine Yaman. "EMG signal classification using discrete wavelet transform and rotation forest." International Conference on Medical and Biological Engineering. Springer, Cham, 2019.

\bibitem{c5} Krizhevsky, A., Sutskever, I., \& Hinton, G. E. (2012). ImageNet Classification with Deep Convolutional Neural Networks. Advances in Neural Information Processing Systems 25.

\bibitem{c6} Ioffe, S., \& Szegedy, C. (2015). Batch Normalization: Accelerating Deep Network Training by Reducing Internal Covariate Shift. Proceedings of the 32nd International Conference on International Conference on Machine Learning, (pp. 448–456).


\bibitem{c7} Srivastava, N., Hinton, G., Krizhevsky, A., Sutskever, I., \& Salakhutdinov, R. (2014). Dropout: a simple way to prevent neural networks from overfitting. The Journal of Machine Learning Research, 1929–1958.


%\bibitem{c8} Shimodaira, H. (2000). Improving predictive inference under covariate shift by weighting the log-likelihood function. Journal of Statistical Planning and Inference, 227–244.



\bibitem{Abadi2015} Abadi, Martın, et al. "TensorFlow: Large-scale machine learning on heterogeneous systems, software available from tensorflow.org (2015)." URL https://www.tensorflow.org (2015).


\bibitem{Chollet2015} Chollet, F. \& others, 2015. Keras. Available at: https://github.com/fchollet/keras.

\end{thebibliography}
\end{document}